\documentclass[conference]{IEEEtran}
\IEEEoverridecommandlockouts
\usepackage{cite}
\usepackage{amsmath,amssymb,amsfonts}
\allowdisplaybreaks
\usepackage{algorithmic}
\usepackage{graphicx}
\usepackage{textcomp}
\usepackage{pifont}
\usepackage{dsfont}
\usepackage{booktabs}
\usepackage{colortbl}
\usepackage[table,xcdraw]{xcolor}
\usepackage{multirow}
\usepackage{marvosym}
\def\BibTeX{{\rm B\kern-.05em{\sc i\kern-.025em b}\kern-.08em
    T\kern-.1667em\lower.7ex\hbox{E}\kern-.125emX}}

\newcommand{\cm}{\ding{52}}

\begin{document}

\title{Let Video Teaches You More: Video-to-Image Knowledge Distillation using DEtection TRansformer for Medical Video Lesion Detection}

\author{
    \IEEEauthorblockN{1\textsuperscript{st} Yuncheng Jiang $\dagger$\thanks{$\dagger$~Equal contributions.}}
    \IEEEauthorblockA{
        \textit{FNii, CUHK-Shenzhen} \\
        \textit{SSE, CUHK-Shenzhen}\\
        \textit{SRIBD, CUHK-Shenzhen}\\       
        Shenzhen, China \\
        yunchengjiang@link.cuhk.edu.cn
    }
    \and
    \IEEEauthorblockN{2\textsuperscript{nd} Zixun Zhang$\dagger$}
    \IEEEauthorblockA{
        \textit{FNii, CUHK-Shenzhen} \\
        \textit{SSE, CUHK-Shenzhen}\\
        Shenzhen, China \\
        zhangchlr1995@gmail.com
    }    
    \and
    \IEEEauthorblockN{3\textsuperscript{rd} Jun Wei$\dagger$}
    \IEEEauthorblockA{
        \textit{FNii, CUHK-Shenzhen} \\
        \textit{SSE, CUHK-Shenzhen}\\
        Shenzhen, China \\
        junwei@link.cuhk.edu.cn
    }
    \and
    \IEEEauthorblockN{4\textsuperscript{th} Chun-Mei Feng, }
    \IEEEauthorblockA{
        \textit{IHPC, A*STAR} \\
        Singapore \\
        strawberry.feng0304@gmail.com    
        } 
    \and
    \IEEEauthorblockN{5\textsuperscript{th} Guanbin Li, }
    \IEEEauthorblockA{
        \textit{SRIDB, CUHK-Shenzhen} \\
        Shenzhen, China \\
        liguanbin@cuhk.edu.cn
    }    
    \and
    \IEEEauthorblockN{6\textsuperscript{th} Xiang Wan}
    \IEEEauthorblockA{
        \textit{SRIBD, CUHK-Shenzhen}\\
        Shenzhen, China \\
        wanxiang@sribd.cn
    }
    \and    
    \IEEEauthorblockN{7\textsuperscript{th} Shuguang Cui}
    \IEEEauthorblockA{
    \textit{SSE, CUHK-Shenzhen}\\
    \textit{FNii, CUHK-Shenzhen} \\
    Shenzhen, China \\
    shuguangcui@cuhk.edu.cn
    }
    \and 
    \IEEEauthorblockN{8\textsuperscript{th} Zhen Li\textsuperscript{\Letter}\thanks{\textsuperscript{\Letter}~Corresponding authors}}
    \IEEEauthorblockA{
        \textit{SSE, CUHK-Shenzhen}\\
        \textit{FNii, CUHK-Shenzhen} \\
        Shenzhen, China \\
        lizhen@cuhk.edu.cn
    }
    \and
}

\maketitle

\begin{abstract}
AI-assisted lesion detection models play a crucial role in the early screening of cancer. However, previous image-based models ignore the inter-frame contextual information present in videos. On the other hand, video-based models capture the inter-frame context but are computationally expensive. 
To mitigate this contradiction, we delve into \textbf{V}ideo-to-\textbf{I}mage knowledge distillation leveraging \textbf{DE}tection \textbf{TR}ansformer (\textbf{V2I-DETR}) for the task of medical video lesion detection. 
V2I-DETR adopts a teacher-student network paradigm. The teacher network aims at extracting temporal contexts from multiple frames and transferring them to the student network, and the student network is an image-based model dedicated to fast prediction in inference. By distilling multi-frame contexts into a single frame, the proposed V2I-DETR combines the advantages of utilizing temporal contexts from video-based models and the inference speed of image-based models. Through extensive experiments, V2I-DETR outperforms previous state-of-the-art methods by a large margin while achieving the real-time inference speed (\textit{30} FPS) as the image-based model.
\end{abstract}


\section{Introduction}
\label{sec:introduction}
\begin{figure}[t]
    \centering
    \includegraphics[width=\linewidth]{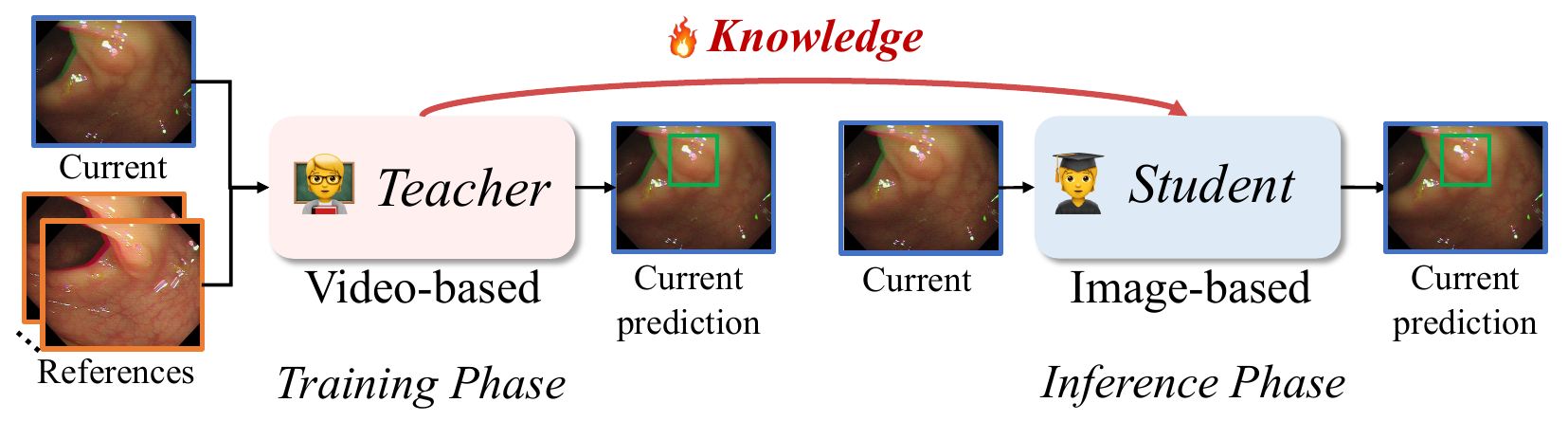}
    \caption{\textbf{Motivation of video-to-image knowledge transfer.} In the training phase, a novel knowledge distillation with multi-frame collaboration is designed to transfer the rich temporal information to the student. We note the student network only takes a single image as input during inference.}
    \label{fig:teaser}
\end{figure}

\begin{figure*}[t]
\centering
\includegraphics[width=\linewidth]{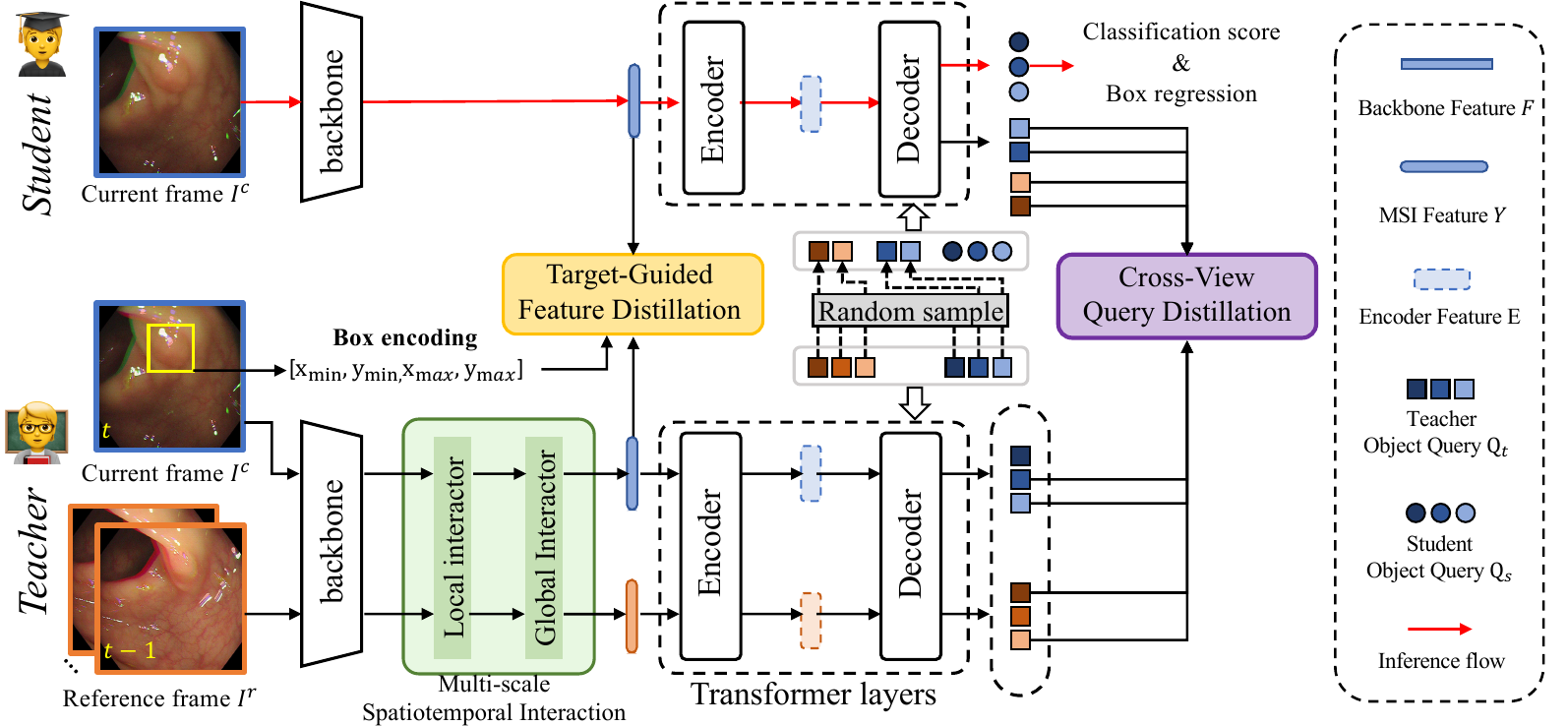}
\caption{\textbf{Pipeline of V2I-DETR}. 
    In the training phase, a Multi-scale Spatiotemporal Interaction module (MSI) is designed to enable the interaction of temporal contexts from reference frames. To guide the student model, we adaptively transfer the foreground features from the teacher to the student by Target-Guided Feature Distillation (TFD) and transfer the temporal relations of target proposals to the student Decoder by Cross-View Query Distillation (CQD).
    During inference, only the student model is employed to conduct image-level prediction.
    Figure best viewed in color.}
\label{fig:framework}
\end{figure*}

Cancer has emerged as a significant global health concern, resulting in tens of millions of deaths annually~\cite{siegel2023cancer}. To prevent cancer at an early stage, AI-based screening techniques have been extensively employed~\cite{baxter2009association} to assist radiologists in performing rapid and accurate diagnoses~\cite{siegel2023cancer}.
Especially in recent years, \textbf{V}ideo-based \textbf{L}esion \textbf{D}etection (VLD) has emerged as a popular research topic that has garnered significant attention.

VLD encompasses two primary paradigms: image-based and video-based methodologies.
For image-based methods~\cite{movahedi2020automated, pacal2022efficient, jiang2023ecc}, video clips are disassembled into separate images. Notably, DEtection TRansformer (DETR)~\cite{carion2020end} stands out as one of the most representative methods. It significantly simplifies the detection pipeline and achieves dominant performance on the detection task. Despite its great potential, those methods neglected inter-frame temporal contexts as each image is processed individually. Consequently, image-based methods fail to handle challenging scenarios, such as distortions, blurriness, occlusions, and motion artifacts in videos.
In contrast, video-based detection models~\cite{wu2021multi,jiang2023yona,lin2022new} excel at extracting temporal contexts, making them robust to video disturbances and outperforming image-based detectors. 
Wu \textit{et al.} \cite{wu2021multi} first considers temporal and spatial attention in multi-frame collaboration.
Jiang \textit{et al.} \cite{jiang2023yona} further introduces efficient tempospatial alignment using only one adjacent reference frame.
Lin \textit{et al.} \cite{lin2022new} first collects and annotates an ultrasound video dataset for breast lesion detection.
However, these works use post-process methods or aggregate too many frames, leading to slow inference speed.However, these methods are computationally expensive, as they need to process multiple frames ({\it i.e.}, current frame, and reference frames) simultaneously, thereby increasing the computational burdens and rendering them impractical for clinical scenarios mandating real-time performance.

To address the above issues, we propose the \textit{first} \textbf{V}ideo-to-\textbf{I}mage knowledge distillation framework leveraging \textbf{DE}tection \textbf{TR}ansformer~\cite{he2021end} (named \textbf{V2I-DETR}), which strikes a balance between speed and accuracy for the task of medical video lesion detection. Fig.~\ref{fig:teaser} shows our main idea which consists of a teacher network and a student network. During training, the teacher network (video-based model) extracts spatial and temporal contexts from multiple frames and transfers the knowledge to the student network (image-based model). During inference, only the student network is utilized.
To extract inter-frame temporal contexts in the teacher network, we introduce the \textbf{Multi-scale Spatiotemporal Interaction} (MSI) module. 
MSI combines the ability of convolution and self-attention to enhance the teacher network's capability for integrating inter-frame temporal contexts.
Additionally, we transfer the knowledge from teacher to the student in two aspects:
1) \textit{At the backbone feature level}. Features from the backbone encoder contain substantial contexts, thus we design the \textbf{Target-guided Feature distillation} (TFD) module to emphasize the target-specific details by fully harnessing the bounding box as a prior which effectively suppresses the irrelevant background noise.
2) \textit{At the decoder query level}. 
The object query is essential for identifying object locations and sizes as it retains detailed object information. To harness this potential, we introduce the \textbf{Cross-view Query Distillation} (CQD) module to reduces background noise and brings the regression targets closer to the ground truth.
Extensive experiments evaluated on SUN-SEG \cite{sunwebsite} and BUV \cite{lin2022new} datasets demonstrate the effectiveness of our proposed V2I-DETR.

In summary, our main contributions are threefold:  
\begin{itemize}
    \item We propose a DETR-based video lesion detection model based on Teacher-Student distillation architecture, called V2I-DETR. To the best of our knowledge, it is the first DETR-based detector with multi-frame collaboration on the VLD task. 
    \item We propose a MSI in the teacher model to extract temporal contexts at different scales and transfer the knowledge to the student model by TFD and CQD, thereby eliminating computation burdens for inference while achieving superior detection performance.
    \item Extensive experiments show that our V2I-DETR outperforms previous state-of-the-art image-based/video-based methods by a large margin on both colonoscopy and breast ultrasound datasets.
\end{itemize}

\section{Method}
\label{method}
\subsection{Overview}  
The pipeline for V2I-DETR is illustrated in Fig.~\ref{fig:framework}, which is developed upon a teacher-student architecture.
Both the teacher and student are implemented using the same DEtection TRansformer model.
During training, the teacher model takes multiple images as input, including current frame $I^c \in \mathbb{R}^{C \times H\times W}$ and several reference frames $I^r \in \mathbb{R}^{C\times (T-1) \times H \times W}$. Through the Multi-scale Spatiotemporal Interaction (MSI) module, the teacher model adeptly captures both spatial and temporal contexts from these diverse frames. Subsequently, these learned contexts are effectively transferred to the student model. This transfer is facilitated by techniques like Target-guided Feature Distillation (TFD) and Cross-view Query Distillation (CQD), which contribute to enhancing the student model's understanding and performance.
During inference, the teacher model is disregarded, and only the student model is utilized for making predictions, which takes a single image (current frame $I^c$) as input, where $C,T,H,W$ are the channel, frame number, image height, and width, respectively.
Further specifics of V2I-DETR will be discussed in the following sections.

\subsection{Multi-scale Spatiotemporal Interaction}
\label{ssec:temporal_fusion}
\begin{figure}[t]
    \centering
    \includegraphics[width=\linewidth]{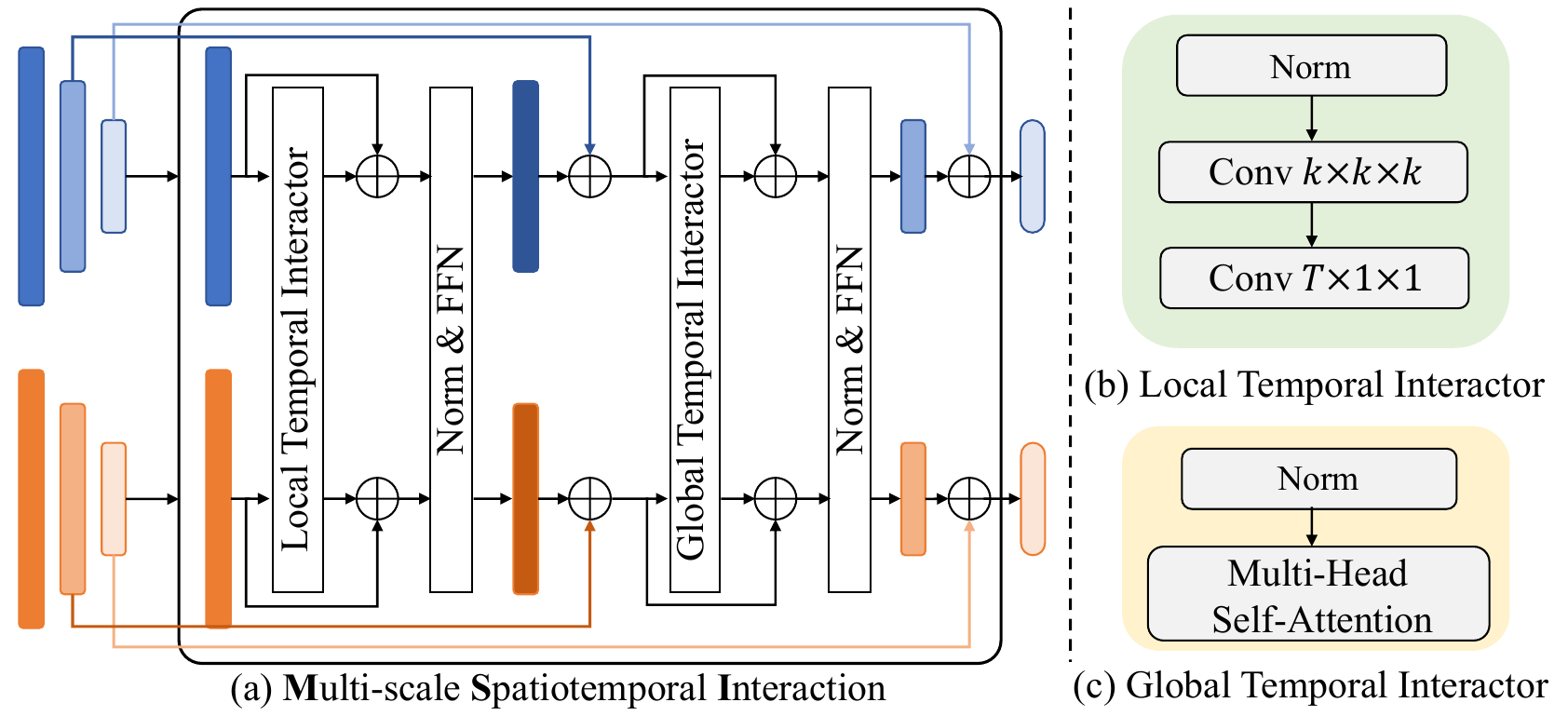}
    \caption{\textbf{Multi-scale Spatiotemporal Interaction (MSI) Module}. Backbone features from different frames interact at each scale, where low-level features are first enhanced by local attention while high-level features are further aggregated by global attention. $k$ is the kernel size.}
    \label{fig:temporal_fusion}
\end{figure}

As indicated by \cite{li2021uniformer}, video inherits a high degree of contextual redundancy and dependency. 
Specifically, low-level features typically rely on local rather than global contexts to learn object relations, but high-level features require global features to build long-range dependencies across distant frames.
Motivated by this, we propose the MSI module to enable low-level/high-level feature interaction between different scales of multiple frames. 
As shown in Fig.~\ref{fig:temporal_fusion}, the MSI comprises two interaction layers, each of which consists of a temporal interact block $\mathcal{A}$ and a Feed-Forward Network (FFN).
Given the multi-scale teacher features $F_t^l \in \mathbb{R}^{d\times T\times h\times w},~l=0,1,2,3$, where $h=\frac{H}{2^l},w=\frac{W}{2^l}$ are the height and width of feature at different level $l$. 
The enhanced feature $Y_t^l$ at each level can be obtained: 

\begin{equation}
    \begin{aligned}
    Y_t^l & = \text{FFN}(\mathcal{A}^l(\text{PE}(F_t^l)) + F_t^l),\\
\end{aligned}
\end{equation}
where PE is the patch embedding to integrate 3D position information into tokens, and $\mathcal{A}^l$ is the interact block of level $l$.
In the shallow layers ($l=0,1$), we extract the spatial and temporal affinity between tokens in small volumetric regions by local interactor using the 3D convolutional layer to encourage the network to capture contextual information between adjacent frames.
In the deep layers ($l=2,3$), we focus on capturing long-term token dependency by global interactor.
We compute the feature similarity of all tokens by self-attention mechanisms~\cite{vaswani2017attention}. 
Hence, our model exhibits insensitivity to the changes in the size and location of moving objects.

\subsection{Target-guided Feature Distillation}
\label{ssec:feature_distill}
The detection performance is greatly influenced by the feature representations of the backbone, primarily due to their abundance of semantic information associated with objects.
Therefore, it is necessary to mimic the spatiotemporal features of the teacher model to boost the performance of the student model.
An intuitive manner to transfer the teacher's knowledge to the student is reducing the Euclidean distance between feature $F^3_s$ and $Y^3_t$~\cite{romero2014fitnets}:
\begin{equation}
    \mathcal{L}_{bk} = \frac{1}{N}\sum^w_{i=1}\sum^h_{j=1}\sum^d_{k=1}|(\text{F}^3_s)^{ijk} - (F^3_t)^{ijk}|
\end{equation}
where $N$ is the total pixel number, $(\text{F}^3_s)$ and $(F^3_t)$ denote the last-stage feature of the current frame from the student and teacher backbone, respectively. 
However, foreground objects are relatively smaller compared to backgrounds. 
If features from the teacher model and the student model are directly aligned, foreground features might be overshadowed and a subsequent drop in performance.

Hence, we employ the ground truth bounding boxes as indicators to guide the model to focus more on the foreground portions of each image.
Furthermore, instead of adopting a rigid hard selection method like~\cite{sun2020distilling}, which exclusively emphasizes foreground pixels within the ground truth box for distillation points, we opt for a more adaptable approach by employing soft selection since the background information also holds valuable insights and contributes to improved model generalization.
Specifically, given a bounding box $B$ of an object, with the size of $H \times W$ and centered at $\Tilde{x}, \Tilde{y}$, we splat all ground truth points onto a two-dimensional Gaussian heatmap $H=exp(-\frac{(x-\Tilde{x})^2 + (y-\Tilde{y})^2}{2\sigma_p^2})$,
where $(x,y) \in B$ and the $\sigma_p$ is object size-adaptive factor along two directions. 
The final soft selection mask is obtained by normalization $M = \frac{H}{||H||_2}$.
In particular, if several Gaussian masks overlap over a single pixel $(x, y)$, we take the element-wise maximum. 
Fig.~\ref{fig:visual}a illustrates some images and the corresponding Gaussian masks.
The soft selection mask is only effective within the activated Gaussian mask, thus only useful information is transferred to the student model.
With the delicately designed Gaussian mask, the backbone features are distilled via minimizing the following loss:

\begin{equation}
    \label{eq:l_bk}
    \mathcal{L}_{bk} = \frac{1}{N}\sum^w_{i=1}\sum^h_{j=1}\sum^d_{k=1}M^{ijk}|(F^3_s)^{ijk} - (Y^3_t)^{ijk}|.
\end{equation}

\begin{table*}[t]
    \centering
    \caption{Quantitative comparison of different detectors and our V2I-DETR on SUN Dataset with ResNet-50 backbone. I: image-based model. V: video-based model. $\rightarrow$: KD manner.}
    \resizebox{0.9\linewidth}{!}
    {\begin{tabular}{lccccccccc}
    \toprule[0.1pt]
    \multirow{2}{*}{Models} & \multirow{2}{*}{Manner} & \multicolumn{3}{c}{SUN Easy Test} &\multicolumn{3}{c}{SUN Hard Test} & \multirow{2}{*}{FPS} & \multirow{2}{*}{Params (M)}\\ \cline{3-8} 
    & & Precision & Recall & F1-score & Precision & Recall & F1-score \\ 
    \midrule
    FGFA & V & 80.2 &  72.1 & 75.9 &  78.9 &  70.4 &  74.4 &   5 & 29  \\
    MEGA & V & 82.0 &  75.6 & 78.7 &  80.4 &  71.6 &  75.7 &   8 &40 \\
    STFT & V & 83.5 &  74.8 & 78.9 &  81.5 &  72.4 &  76.7 &  8  &45 \\
    TransVOD & V & 81.5 &  74.0 & 77.6 &  79.3 &  69.6 &  74.1 &   13 & 52 \\
    YONA & V & 85.4 & 75.3 & 80.0 & 83.3 &  74.6 & 78.9 &    \textbf{45} & 25    \\ 
    CVA-Net & V & 86.5 & 76.5 & 81.2 & 87.1 & 75.2 & 80.7 & 10 & 46\\
    \midrule
    Deformable DETR  & \multirow{2}{*}{V} & \multirow{2}{*}{88.5} & \multirow{2}{*}{81.0} & \multirow{2}{*}{84.6} & \multirow{2}{*}{91.5} &  \multirow{2}{*}{76.5} & \multirow{2}{*}{83.3} & \multirow{2}{*}{2} & \multirow{2}{*}{40}\\
    ~~~~+ MSI (Teacher) \\
    \midrule
    DETR & I & 78.4 & 69.8 &  73.9 & 82.4 &  63.5 & 71.7 & 30 & 41 \\
    \rowcolor{gray!30}
    ~~~~+ V2I-DETR & V $\rightarrow$ I & 85.6  & 77.8  &  81.5 & 84.7  & 73.8 & 78.9  & 30 & 51\\
    Deformable DETR & I & 80.0 & 71.3 &  75.4 & 83.6 & 67.1 & 74.4 & 18 & 40 \\
    \rowcolor{gray!30}
    ~~~~+ V2I-DETR & V $\rightarrow$ I & 88.9 & 80.3 & 84.4 & 91.2 &  76.3 & 83.1 & 18 & 50 \\
    SAM DETR & I & 81.0 & 72.5 & 76.5 &  80.4 &  71.2 & 75.5 &  16 & 58\\
    \rowcolor{gray!30}
    ~~~~+ V2I-DETR & V $\rightarrow$ I & \textbf{89.4} & \textbf{82.0} & \textbf{85.5} & \textbf{92.4} &  \textbf{77.5} & \textbf{84.3} & 16 & 68\\
    \bottomrule[1pt]
    \end{tabular}
    }
    \label{tab:polyp_results}
\end{table*}

\begin{table}[t]
    \centering
    \caption{Quantitative comparison of different detectors and our V2I-DETR on BUV Dataset with ResNet-50 backbone.}
    \resizebox{1\linewidth}{!}
    {\begin{tabular}{lcccc}
    \toprule[0.75pt]
    Model & Manner & AP & $\text{AP}_{50}$ & $\text{AP}_{75}$ \\ 
    \midrule
    FGFA & V & 26.1 & 49.7 & 27.0 \\
    MEGA & V & 32.3 &  57.2 & 35.7 \\
    STFT& V & 32.8 & 59.7 & 36.0 \\
    TransVOD & V & 30.4 &  60.6 & 35.6 \\
    YONA & V & 33.6 & 61.7 & 36.5 \\
    CVA-Net & V & 36.1 & \textbf{65.1} & 38.5 \\
    \midrule
    Deformable DETR  & \multirow{2}{*}{V} & \multirow{2}{*}{37.0} &  \multirow{2}{*}{64.2} & \multirow{2}{*}{39.5} \\
    ~~~~+ MSI (Teacher) \\
    \midrule
    DETR & I & 27.6  & 51.2  & 30.2  \\
    \rowcolor{gray!30}
    ~~~~+ V2I-DETR & V $\rightarrow$ I & 31.2  & 55.8  & 34.6  \\
    Deformable DETR & I & 31.5 & 54.3 & 33.8 \\
    \rowcolor{gray!30}
    ~~~~+ V2I-DETR & V $\rightarrow$ I & 36.8 & 63.9 & 39.7 \\
    SAM DETR & I & 32.8 & 57.0 & 35.1 \\
    \rowcolor{gray!30}
    ~~~~+ V2I-DETR & V $\rightarrow$ I & \textbf{37.4} & 64.9 & \textbf{40.2} \\
    \bottomrule[0.75pt]
    \end{tabular}}
    \label{tab:breast_results}
\end{table}

\subsection{Cross-View Query Distillation}
\label{ssec:query_distill}
Based on previous research~\cite{zhu2020deformable}, well-optimized queries in the teacher model consistently achieve a stable bipartite assignment and improve the training stability of the student model.
Thus, previous image-to-image knowledge distillation methods for DETR directly apply consistency regularization between object queries.
However, in the multi-frame collaboration paradigm, the information in the current frame and the reference frame is not completely consistent due to the camera moving.

To overcome this issue, we propose cross-view query distillation that enables the DETR-based framework to learn semantically invariant characteristics of object queries from two different views.
The details of the cross-view query distillation process are shown in Fig.~\ref{fig:framework}.
Specifically, for object query $\text{Q}_t \in \mathbb{R}^{N\times d}$ of each frame in the teacher model, we randomly select $n$ queries $q_t \in \mathbb{R}^{n\times d} = \{\text{Q}_t^i,~i \in \Phi(N)\}$ from $\text{Q}_t$ as the cross-view query embeddings,
where $n = \lfloor N/T\rfloor$ and $\Phi(\cdot)$ is a random selection function that returns a sequence of list index of the teacher object query. 
We propose the random selection strategy to regulate the number of cross-view queries to be roughly equal to the number of student queries, thus student queries will not be overwhelmed by larger cross-view queries. 
Then the cross-view query embedding is attached to the original student object queries in another view to serve as the input of the decoder:

\begin{equation}
    \begin{aligned}
        \hat{D}_s, D^c_s &= \text{Decoder}_s([q_t, \text{Q}^c_s], \text{E}_s), \\
        D^r_t, D^c_t     &= \text{Decoder}_t([\text{Q}^r_t, \text{Q}^c_t], \text{E}_t).
    \end{aligned}
\end{equation}

\noindent Where $\hat{D}$ and $D$ denote the decoded features of cross-view queries and original object queries, and $E$ denotes the encoded image features.   
Note that subscripts $t$ and $s$ indicate teacher and student, respectively. 
With the semantic guide of input cross-view query embeddings, the correspondence of the decoded features can be naturally guaranteed, and we impose consistency loss as follows:

\begin{equation}
    \label{eq:l_qe}
    \mathcal{L}_{qe} = -\frac{\hat{D}\cdot D}{||\hat{D}||_2\times||D||_2}.
\end{equation}

\subsection{Loss Function}
\label{ssec:loss}
To sum up, the total loss for training the V2I-DETR is a weighted combination of bounding box supervision and knowledge distillation:
\begin{equation}
    \mathcal{L} = \mathcal{L}_{det} + \lambda_{bk}\mathcal{L}_{bk} + \lambda_{qe}\mathcal{L}_{qe},
\end{equation}
where $\mathcal{L}_{det}$ is the sum of f1 loss and giou loss as described in~\cite{carion2020end}, $\lambda_{bk}$, and $\lambda_{qe}$ are the weights for each individual distillation loss. 
To guarantee the loss terms are roughly of the same magnitude, we fine-tune weights with various combinations as shown in Fig.~\ref{fig:ablation}(a), and empirically fix those to $\lambda_{bk} = 1$, and $\lambda_{qe} = 5$ in the experiments.
\begin{figure*}[t]
    \centering
    \includegraphics[width=1\linewidth]{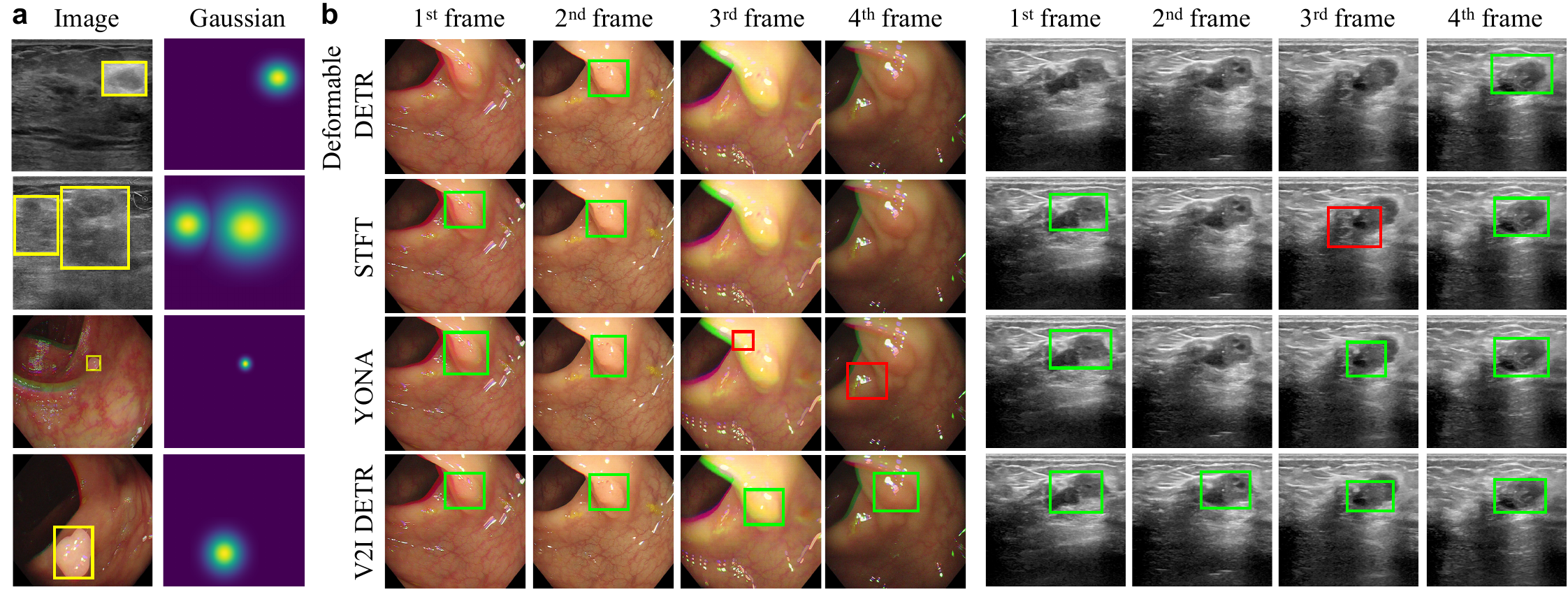}
    \caption{\textbf{Qualitative comparisons}. Visualization results produced by our method and state-of-the-art methods on SUN video dataset (left part) and BUV dataset (right part). The \textcolor{green}{green} box indicates the correct predictions and the \textcolor{red}{red} box denotes the wrong predictions.}
    \label{fig:visual}
\end{figure*}


\section{Experiment}
\label{sec:experiment}

\textbf{Datasets}  
We evaluate the proposed method on two large-scale lesion detection benchmarks corresponding to two clinical tasks: 1) SUN Colonoscopy Video (SUN)~\cite{suncolon}: 112 clips are used for training and 54 clips for testing.
2) Breast Ultrasound Video (BUV)~\cite{lin2022new}: 150 videos were used as a training set, and 38 videos were used as the testing set.

\textbf{Implementation Details} 
We leverage Deformable DETR with the proposed MSI module as the teacher model and use three DETR-like detectors (DETR~\cite{carion2020end}, Deformable DETR~\cite{zhu2020deformable}, and SAM-DETR~\cite{zhang2022accelerating}) as student models for evaluation. 
Meanwhile, five state-of-the-art video object detectors are used for fair comparison, including FGFA~\cite{zhu2017flow}, MEGA~\cite{chen2020memory}, STFT~\cite{wu2021multi}, TransVOD~\cite{he2021end}, YONA~\cite{jiang2023yona}, and CVA-Net~\cite{lin2022new}.
We use ResNet-50~\cite{he2016deep} pre-trained on ImageNet as the backbone.
We set the number of object queries $N=100$ and training frames of the video-based model $T=3$.
Our network is implemented on Pytorch and trained using AdamW with an initial learning rate $10^{-4}$ and batch size $16$ for $12$ epochs. For the colonoscopy images, we take the precision, recall, and f1-scores following~\cite{bernal2017comparative} to evaluate the performance. 
For the Breast images, we follow~\cite{lin2022new} and use average precision $\text{AP}$, $\text{AP}_{50}$, and $\text{AP}_{75}$ for evaluation.

\subsection{Comparison with SOTA methods}
The results of colonoscopy lesion detection are displayed in Tab.~\ref{tab:polyp_results}.  
Primarily, the video-based teacher model, utilizing multiple frames as inputs, exhibits a significant performance advantage over the original Deformable DETR by $9.2\%$ and $7.6\%$ f1-score on the easy and hard test set. 
However, The cost of performance improvement is very low inference frames per second (FPS).
In contrast, our V2I-DETR model, employing knowledge distillation, consistently enhances the performance of the student model. It yields improvements of f1-score by $7.2\% \sim 9.0\%$ on both test sets.  
Specifically, V2I-DETR based on SAM DETR achieves the state-of-the-art (SOTA) results, attaining an f1-score of $85.5\%$ (easy) and $84.3\%$ (hard). 
Notably, V2I-DETR will not introduce any computational overhead into the inference stage and achieve $\times 7$ faster FPS.
Even compared with video-based models, our V2I-DETR based on SAM DETR surpasses the SOTA method YONA by $4.3\%$ (easy) and $3.6\%$ (hard) f1-score.
Tab.~\ref{tab:breast_results} demonstrates the comparison on the BUV benchmark. 
Specifically, among the comparison methods, the video-based CVA-Net achieves the highest performance of 36.1\% $\text{AP}$, 65.1\% $\text{AP}_{50}$, 38.5\% $\text{AP}_{75}$.  
Compared to CVA-Net, our V2I-DETR based on Deformable DETR improves the $\text{AP}$ by 0.7\% and $\text{AP}_{75}$ by 1.2\%. While our V2I-DETR based on SAM DETR further boosts the performance by 1.3\% $\text{AP}$ and 1.7\% $\text{AP}_{75}$.

\begin{table}[t]
    \centering
    \caption{Ablation of component effectiveness. TFD: Target-guided Feature Distillation. MSI: Multi-Scale Spatiotemporal Interaction. CQD: Cross-view Query Distillation.}
    \resizebox{\linewidth}{!}{\begin{tabular}{ccccccc}
    \toprule[0.75pt]
    ID & TFD  & MSI & CQD  & Precision & Recall & F1 score   \\ 
    \midrule
    1  & \multicolumn{3}{c}{baseline} &   83.6    &   67.1 &   74.4   \\
    2  & \cm  &     &                 &   85.1$_{\textcolor{green}{\uparrow}  1.5}$    &   72.5$_{\textcolor{green}{\uparrow} 5.4}$ &   78.3$_{\textcolor{green}{\uparrow}  3.8}$   \\
    3  & \cm  & \cm &                 &   88.3$_{\textcolor{green}{\uparrow}  4.7}$    &   74.5$_{\textcolor{green}{\uparrow} 7.4}$ &   80.8$_{\textcolor{green}{\uparrow}  6.4}$   \\
    4  & \cm  &     & \cm             &   89.5$_{\textcolor{green}{\uparrow}  5.9}$    &   75.2$_{\textcolor{green}{\uparrow} 8.1}$ &   81.7$_{\textcolor{green}{\uparrow}  7.3}$   \\
    5  & \cm  & \cm & \cm             &   91.2$_{\textcolor{green}{\uparrow}  7.6}$    &   76.3$_{\textcolor{green}{\uparrow} 9.2}$ &   83.1$_{\textcolor{green}{\uparrow}  8.6}$   \\
    \bottomrule[0.75pt]
    \end{tabular}
    }
    \label{tab:ablation_componemt}
\end{table}

\begin{table}[t]
    \centering
    \caption{Ablation study of distillation methods under different designs. FD: (backbone) Feature Distillation. ED: (encoder) Embedding Distillation. QD: (decoder) Query Distillation. cq:number of cross-view query. sq: number of student query. T: number of frames.}
    \resizebox{\linewidth}{!}{\begin{tabular}{ccccc}
    \toprule[0.75pt]
    ID &  Design                       & Precision  & Recall & F1 score  \\ 
    \midrule
    1  &  Deformable DETR (baseline)   &  83.6     & 67.1   & 74.4      \\
    2  &  w/ FD                        &  83.2$_{\textcolor{red}{\downarrow}  0.4}$       & 70.2$_{\textcolor{green}{\uparrow}  3.1}$   & 76.1$_{\textcolor{green}{\uparrow}  1.7}$      \\
    3  &  w/ FD (GT mask)                 &  84.2$_{\textcolor{green}{\uparrow}  0.6}$      & 71.0$_{\textcolor{green}{\uparrow}  3.9}$   & 77.0$_{\textcolor{green}{\uparrow}  2.6}$      \\
    4  &  w/ FD (attention mask)    &  83.5$_{\textcolor{red}{\downarrow}  0.1}$      &   68.2$_{\textcolor{green}{\uparrow}  1.1}$     &    75.1$_{\textcolor{green}{\uparrow}  0.6}$     \\
    5  &  w/ TFD                       &  85.1$_{\textcolor{green}{\uparrow}  1.5}$      & 72.5$_{\textcolor{green}{\uparrow}  5.4}$   & 78.3$_{\textcolor{green}{\uparrow}  3.8}$      \\
    \midrule
    6  &  w/ ED                        &  80.1$_{\textcolor{red}{\downarrow}  3.5}$      & 62.9$_{\textcolor{red}{\downarrow}  4.2}$   & 70.5$_{\textcolor{red}{\downarrow}  4.0}$      \\
    7  &  w/ ED (attention mask)   &   83.1$_{\textcolor{red}{\downarrow}  0.5}$     &   66.3$_{\textcolor{red}{\downarrow}  0.8}$     &    73.8$_{\textcolor{red}{\downarrow}  0.7}$     \\
    \midrule
    8  &  w/ QD                        &  82.7$_{\textcolor{red}{\downarrow}  0.9}$      & 69.7$_{\textcolor{green}{\uparrow}  2.6}$   & 75.6$_{\textcolor{green}{\uparrow}  1.2}$      \\
    9  &  w/ CQD ($cq=T*sq$)           &  81.5$_{\textcolor{red}{\downarrow}  2.1}$      & 67.1$_{\textcolor{green}{\uparrow}  0}$   & 73.6$_{\textcolor{red}{\downarrow}  0.8}$      \\
    10 &  w/ CQD ($cq=T*sq/2$)         &  83.1$_{\textcolor{red}{\downarrow}  0.5}$      & 71.4$_{\textcolor{green}{\uparrow}  4.3}$   & 76.8$_{\textcolor{green}{\uparrow}  2.4}$      \\
    11 &  w/ CQD ($cq \approx sq$)     &  84.7$_{\textcolor{green}{\uparrow}  1.1}$      & 73.4$_{\textcolor{green}{\uparrow}  6.3}$   & 78.6$_{\textcolor{green}{\uparrow}  4.2}$      \\
    \bottomrule[0.75pt]
    \end{tabular}}
\label{tab:ablation_distill}
\end{table}

\subsection{Visualization}
Fig.~\ref{fig:visual}b visually compares detection results for polyp and breast lesion detection results produced by our network and four compared methods.  
Lack of temporal consistency information, image-based model Deformable DETR fails to detect the polyp and breast lesions, as shown in the first row of Fig.~\ref{fig:visual}b, 
STFT and YONA obtain a better lesion detection result but also fail to detect lesions or produce many false positive results ($3^{rd}, 4^{th}$ in polyp and $2^{nd}$ in the breast).  
In contrast, our method can correctly detect the lesions of all video frames. 
Confirming that our method is effective in precisely detecting lesion regions from moving-object medical videos.

\subsection{Ablation Study}

\textbf{Effectiveness of component}
We investigate the contribution of each proposed component. The results are shown 
in Tab.~\ref{tab:ablation_componemt}. Compared with the original Deformable DETR (\# 1), our proposed components enjoy consistent performance improvement.   
Specifically, by introducing the TFD (\# 2) for feature distillation, it outperforms the baseline by $3.8\%$ on the f1-score.   
Further integrating the MSI (\# 3) for spatiotemporal feature interaction in the teacher model brings an improvement of $6.4\%$. 
Finally, by combining both feature and query distillation, our V2I-DETR framework w/o (\# 4) or w/ (\# 5) MSI can significantly boost the performance by $7.3\%$ and $8.6\%$ on f1-score. 
These results show that our proposed components are complementary to each other and therefore prove the effectiveness of each component in V2I-DETR.

\textbf{Analysis on distillation feature}
Tab.~\ref{tab:ablation_distill} summarized the results of distillation from different stages of DETR. 
Firstly, we apply direct knowledge transfer with MSE loss to the baseline model (\# 2). 
This design slightly improves the performance by $1.7\%$ f1-score, showing that mimicking the teacher backbone learning during training is helpful to the student model.
However, distilling the feature embedding from the DETR encoder (\# 6) causes an obvious performance drop by $4.0\%$ f1-score.
we further apply the self-attention mechanism introduced in~\cite{chang2022detrdistill} to restrict the region of interest (\# 7), but the performance is still worse than the baseline. 
Lastly, we transfer the teacher decoders' object query to the student, which shows a positive contribution to the performance with $1.2\%$ f1-score promotion.
Therefore, we mainly focus on the design of FD and QD in our model. 

\textbf{Analysis on the region of FD} 
Based on the above observation, we dig deeply with different features distillation (FD) strategies as shown in the first part of Tab.~\ref{tab:ablation_distill}.
We first follow the idea in~\cite{chang2022detrdistill} that computes the attention mask between object queries and backbone features to select the valuable region for distillation (\# 4). 
However, the improvement is minimal compared with vanilla distillation (\# 2) and requires significant additional computation.
Next, we guided the distillation using masks generated from the ground truth boxes, which improved precision by 2.6\% in F1-score. Finally, we converted the hard ground truth masks to soft masks using a Gaussian distribution. This further improved performance by 3.8\%, demonstrating that our method effectively balances the information around the target boundaries.

\begin{figure}[t]
    \centering
    \includegraphics[width=\linewidth]{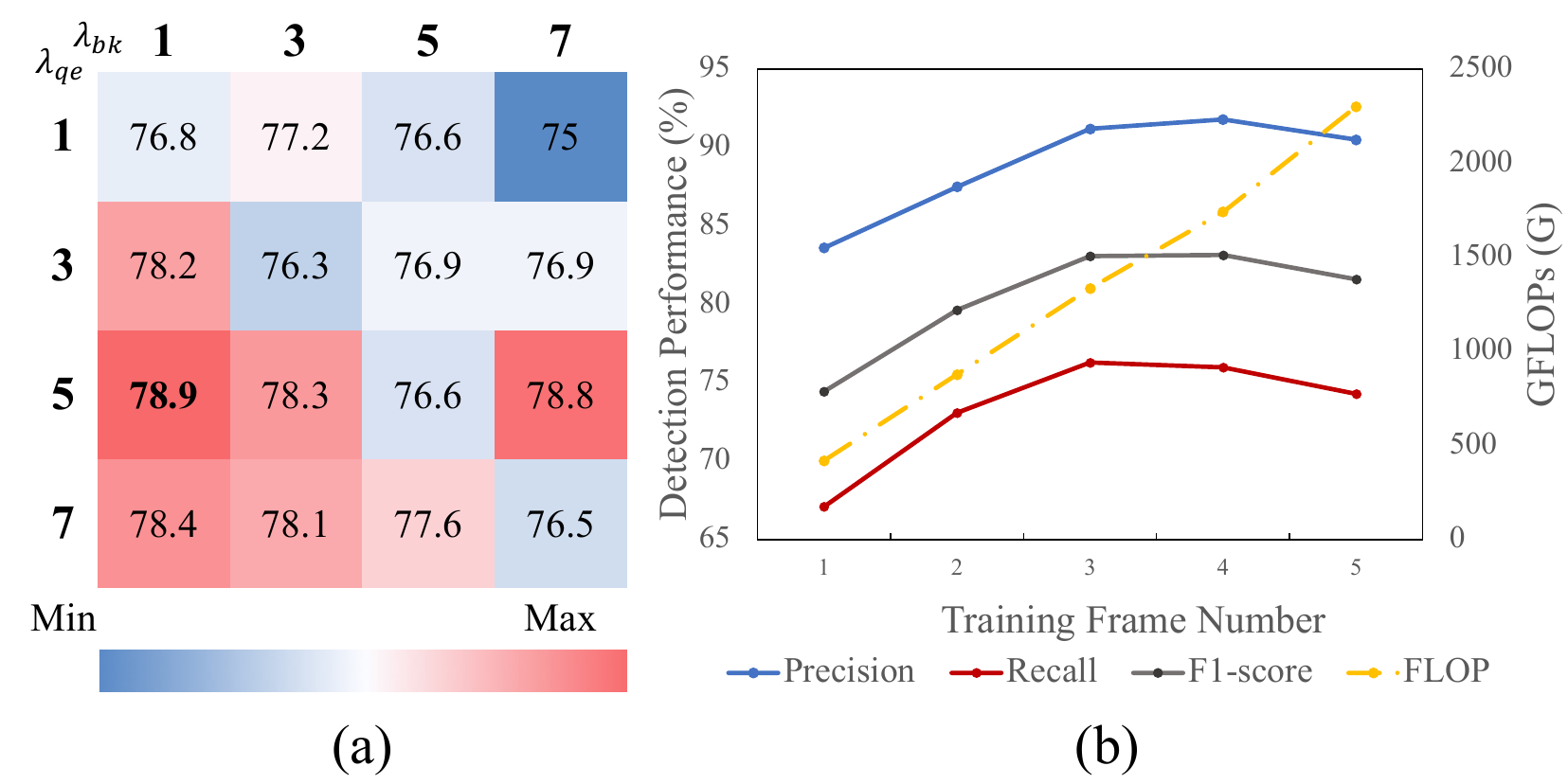}
    \caption{\textbf{Ablation study}. (a) The effect of loss function hyperparameters on detection f1-score. (b) The effect of Training Frame numbers.}
    \label{fig:ablation}
\end{figure}

\textbf{Analysis on the number of CQD}
Meanwhile, we also conduct several ablations on different cross-view numbers for Cross-view Query Distillation (CQD), as shown in The third part of Tab.~\ref{tab:ablation_distill}. 
First, we manually set the number equal to the number of the sum of all teacher queries (\# 9). 
However, because this number was much larger than the number of student queries, it misled the student queries and significantly reduced performance.  
Reducing the ratio could mitigate the undesirable result (\# 10). 
Lastly, we use random sampling to keep the cross-view query number and student one to be approximately same (\# 11), resulting in an improvement of $4.2\%$ on f1-score.

\textbf{Analysis on the number of training reference}
We investigated the impact of different support frame numbers on V2I DETR in Fig.~\ref{fig:ablation}(b). 
We keep the inference frame number as $1$ and change the number of training frame numbers. 
Training with 3 frames achieves the best balance between accuracy and computational complexity (5 frames reach the memory cap).

\section{Conclusion}
\label{sec:conclusion}
In this paper, we analyze the limitations of previous pure image-based and video-based models for video lesion detection considering both accuracy and speed and the challenges of adapting DETR architecture with temporal information. 
We contribute the first transformer-based end-to-end video lesion detector (\textit{i.e.} V2I-DETR).  
It consists of a Multi-scale Spatiotemporal Interaction module that enhances the current feature by interacting with reference features in multiple scales, 
a Target-guided Feature Distillation module that adaptively transfers the valuable foreground information of the teacher model to the student model, 
and a Cross-view Query Distillation module that further transfers the target proposal from multiple views to the student model.  
Notably, all these components are used only in training, incurring no computational cost to inference. 
Extensive experiments demonstrate the superiority of V2I-DETR on both colonoscopy and breast ultrasound lesion detection tasks.

\section*{Acknowledgements}
This work was supported by NSFC with Grant No. 62293482, by the Basic Research Project No. HZQB-KCZYZ-2021067 of Hetao Shenzhen HK S\&T Cooperation Zone, by Shenzhen General Program No. JCYJ20220530143600001, by Shenzhen-Hong Kong Joint Funding No. SGDX20211123112401002, by the Shenzhen Outstanding Talents Training Fund 202002, by Guangdong Research Project No. 2017ZT07X152 and No. 2019CX01X104, by the Guangdong Provincial Key Laboratory of Future Networks of Intelligence (Grant No. 2022B1212010001), by the Guangdong Provincial Key Laboratory of Big Data Computing, \%The Chinese University of Hong Kong, Shenzhen CHUK-Shenzhen, by the NSFC 61931024\&12326610, by Key Area R\&D Program of Guangdong Province with grant No. 2018B030338001, by the Key Area R\&D Program of Guangdong Province with grant No. 2018B030338001, by the Shenzhen Key Laboratory of Big Data and Artificial Intelligence (Grant No. ZDSYS201707251409055), and by Tencent \& Huawei Open Fund.

\bibliographystyle{IEEEtran}
\bibliography{IEEEabrv,bibtex}

\end{document}